\documentclass[sigconf, authorversion]{acmart}

\usepackage{multirow}
\usepackage{siunitx}
\usepackage{makecell}

\AtBeginDocument{%
  }

\copyrightyear{2023} 
\acmYear{2023} 
\setcopyright{acmlicensed}
\acmConference[MM '23]{Proceedings of the 31st ACM International Conference on Multimedia}{October 29-November 3, 2023}{Ottawa, ON, Canada}
\acmBooktitle{Proceedings of the 31st ACM International Conference on Multimedia (MM '23), October 29-November 3, 2023, Ottawa, ON, Canada}
\acmPrice{15.00}
\acmDOI{10.1145/3581783.3611810}
\acmISBN{979-8-4007-0108-5/23/10}

\begin{document}

\title{ZRIGF: An Innovative Multimodal Framework for Zero-Resource Image-Grounded Dialogue Generation}

\author{Bo Zhang}
\orcid{0000-0001-6933-922X}
\affiliation{%
  \institution{Dalian University of Technology}
  \streetaddress{No.2 Linggong Road, Ganjingzi District}
  \city{Dalian}
  \country{China}
  \postcode{116024}
}
\email{zhangbo1998@mail.dlut.edu.cn}

\author{Jian Wang}
\authornote{Corresponding author.}
\affiliation{%
  \institution{Dalian University of Technology}
  \streetaddress{No.2 Linggong Road, Ganjingzi District}
  \city{Dalian}
  \country{China}
  \postcode{116024}
}
\email{wangjian@dlut.edu.cn}

\author{Hui Ma}
\affiliation{%
  \institution{Dalian University of Technology}
  \streetaddress{No.2 Linggong Road, Ganjingzi District}
  \city{Dalian}
  \country{China}
  \postcode{116024}
}
\email{huima@mail.dlut.edu.cn}

\author{Bo Xu}
\affiliation{%
  \institution{Dalian University of Technology}
  \streetaddress{No.2 Linggong Road, Ganjingzi District}
  \city{Dalian}
  \country{China}
  \postcode{116024}
}
\email{xubo@dlut.edu.cn}

\author{Hongfei Lin}
\affiliation{%
  \institution{Dalian University of Technology}
  \streetaddress{No.2 Linggong Road, Ganjingzi District}
  \city{Dalian}
  \country{China}
  \postcode{116024}
}
\email{hflin@dlut.edu.cn}

\renewcommand{\shortauthors}{Zhang et al.}

\begin{abstract}
Image-grounded dialogue systems benefit greatly from integrating visual information, resulting in high-quality response generation. However, current models struggle to effectively utilize such information in zero-resource scenarios, mainly due to the disparity between image and text modalities. To overcome this challenge, we propose an innovative multimodal framework, called ZRIGF, which assimilates image-grounded information for dialogue generation in zero-resource situations. ZRIGF implements a two-stage learning strategy, comprising contrastive pre-training and generative pre-training. Contrastive pre-training includes a text-image matching module that maps images and texts into a unified encoded vector space, along with a text-assisted masked image modeling module that preserves pre-training visual features and fosters further multimodal feature alignment. Generative pre-training employs a multimodal fusion module and an information transfer module to produce insightful responses based on harmonized multimodal representations. Comprehensive experiments conducted on both text-based and image-grounded dialogue datasets demonstrate ZRIGF's efficacy in generating contextually pertinent and informative responses. Furthermore, we adopt a fully zero-resource scenario in the image-grounded dialogue dataset to demonstrate our framework's robust generalization capabilities in novel domains. The code is available at \url{https://github.com/zhangbo-nlp/ZRIGF}.
\end{abstract}


%

\maketitle

\section{Introduction}
Dialogue systems have attracted considerable interest in AI research due to their promising applications in virtual assistants and social chatbots \cite{ref1, ref2}. Although existing models generate fluent responses thanks to advanced neural architectures like Transformer \cite{ref4} and pre-training techniques such as DialoGPT \cite{ref5}, a noticeable gap remains when comparing machine-generated dialogues to human-to-human conversations. A primary factor contributing to this disparity is that, unlike humans who can associate dialogue with background knowledge such as visual and auditory perceptions, machines can only access limited information from superficial text.

Bridging the gap between human and machine communication requires incorporating visual scene information into dialogue generation. Recent research has explored various vision-language dialogue tasks, including visual dialogue \cite{ref6, ref7} and image-grounded dialogue \cite{ref8, ref9, ref10, ref16}. Visual dialogue entails answering questions about the factual content of an image through multi-turn conversations, while image-grounded dialogue aims to generate fitting responses to given images and dialogue contexts through casual conversation. The remarkable success of ChatGPT \cite{ref11} has spurred increased research focus on image-grounded dialogue due to its lack of multimodal capabilities. However, there are still some challenges remaining unresolved.

One of the major challenges in image-grounded dialogue research is the scarcity of large-scale dialogue datasets that are naturally related to images, as obtaining such data from human annotators is costly. Although some crowdsourced datasets have been proposed \cite{ref12, ref13, ref14}, their limited training size hinders the development of dialogue generation models applicable to diverse domains. Consequently, there is a growing interest in improving image-grounded dialogue performance within zero-resource scenarios, where context-image-response triples obtained from crowdsourcing remain unused for training purposes. Previous endeavors, such as \cite{ref9} and \cite{ref16}, have explored zero-resource image-grounded dialogue generation; however, these efforts only focused on utilizing retrieved images to enhance response quality, neglecting the crucial aspect of generalization ability in zero-resource scenarios. To address this issue, we propose a novel approach with strong generalization ability that can effectively adapt and perform well in new domains, even devoid of labeled data for fine-tuning.

\begin{figure}
	\centering
	\includegraphics[scale=0.5]{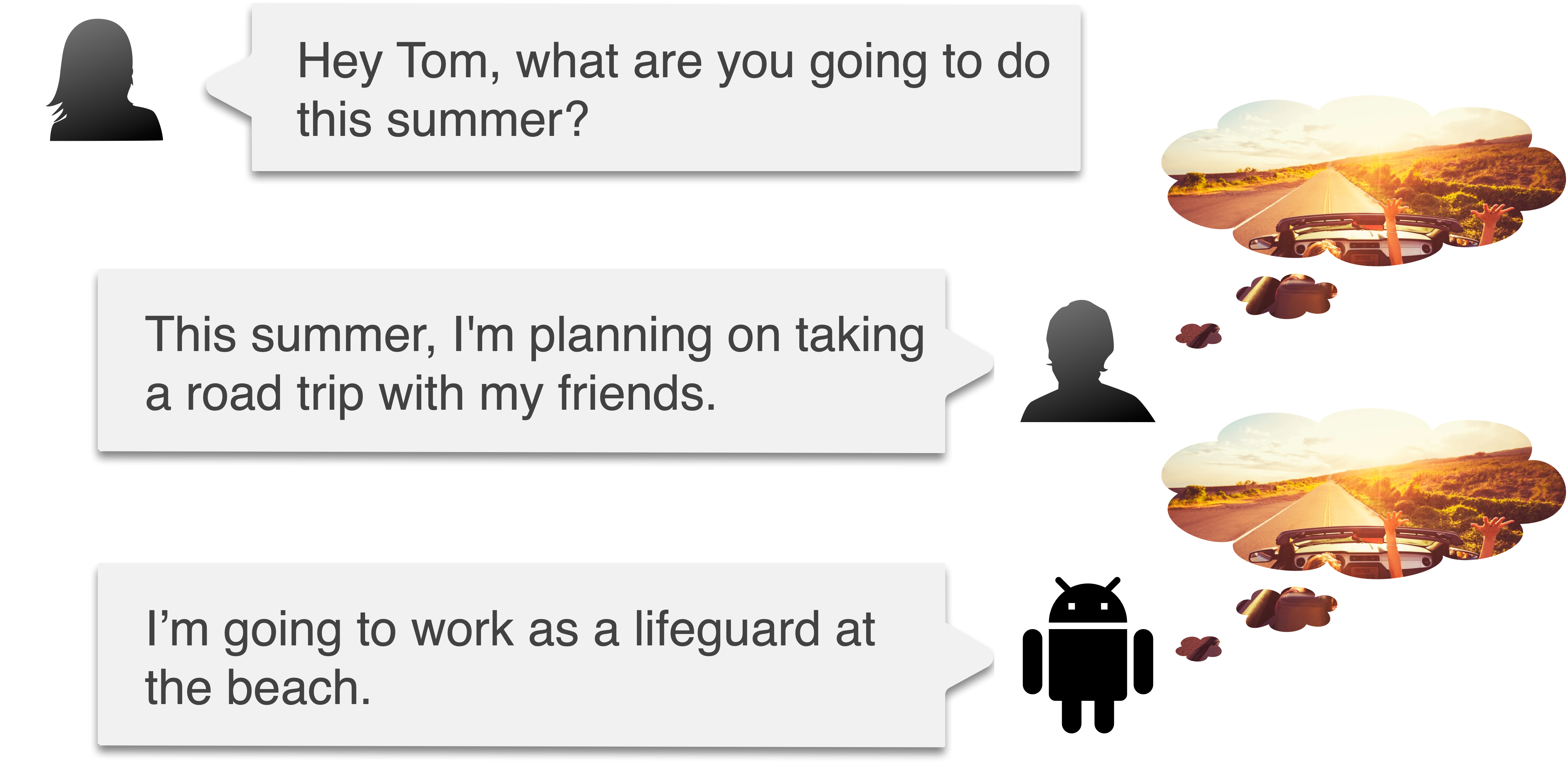}
	\caption{An example of image-grounded dialogue, illustrating a comparison between human and robot responses to a given image.}
	\label{fig_dialog}
\end{figure}

Moreover, a significant modality gap exists between images and texts, particularly between dialogue context and corresponding images, which can result in suboptimal use of visual information. A common approach \cite{ref13} for image-grounded dialogue generation involves encoding images and texts separately and then generating responses during the decoding process; however, this method completely disregards the modality gap. Recent models, such as Maria \cite{ref9} and VisAD \cite{ref16}, have attempted to incorporate visual information into dialogue generation before or during the decoding process. These approaches share similarities with text-only knowledge-grounded dialogue generation \cite{ref17, ref18}, but they fail to completely bridge the modality gap due to the persistent lack of correlation among different modality vectors. As shown in Figure \ref{fig_dialog}, a person asks about summer vacation plans, and the dialogue partner (either a human or a robot) provides a response based on the given image. The human response is coherent and contextually appropriate, while the robot response is mismatched and irrelevant to the image content, which shows a highway and not a beach or any sand. This example shows the importance of aligning different modality vectors for effective image-grounded dialogue generation.

To address the challenges inherent in a fully zero-resource scenario for image-grounded dialogue generation, we present an innovative multimodal encoders-decoder architecture, termed ZRIGF. The encoders and decoder are functionally segregated into four modules: a text-image matching module, a text-assisted masked image modeling module, a multimodal fusion module, and an information transfer module. To combine these modules, we devise a learning strategy consisting of two stages: contrastive pre-training and generative pre-training. \textbf{First}, during the contrastive pre-training stage, the encoders serve as a text-image matching module that maps image and text inputs into a unified encoded vector space, thereby aligning different modality vectors. This matching module trains on a large corpus of annotated image-text pairs using pre-trained image and text encoders. Additionally, we implement a text-assisted masked image modeling module that preserves the original pre-training visual features and realigns the multimodal features. \textbf{Second}, during the generative pre-training stage, a multimodal fusion module and an information transfer module are employed to generate insightful responses based on aligned multimodal representations. These modules further facilitate alignment of representations. The entire model in this stage is trained using a corpus of annotated context-response pairs augmented with relevant images obtained through the text-image matching module.

To assess the efficacy of our proposed ZRIGF, we perform experiments on two publicly available dialogue datasets, namely the text-based Reddit Conversation Corpus \cite{ref19} and the image-grounded Image-Chat dataset \cite{ref12}.
The experimental results demonstrate that ZRIGF outperforms competitive baselines on most automatic evaluation metrics and achieves comparable performance with ChatGPT on human evaluation metrics. Moreover, ZRIGF exhibits remarkable robustness and generalization ability under different data settings, especially in the fully zero-resource scenario.

The contributions of this paper can be summarized as follows:
\begin{itemize}
\item We propose a novel framework for image-grounded dialogue generation within a fully zero-resource scenario, proficiently addressing the modality gap between images and texts through a two-stage learning strategy: contrastive pre-training and generative pre-training.
\item We introduce pioneering components such as a text-assisted masked image modeling module and an information transfer decoder, which further align multimodal representations and generate informative responses.
\item We conduct comprehensive experiments on two publicly available dialogue datasets, showcasing the superior performance and robust generalization ability of our proposed ZRIGF in comparison to state-of-the-art models.
\end{itemize}

\section{Related Work}
\subsection{Image-Grounded Dialogue Generation}
Image-grounded dialogue generation is a task that seeks to generate natural and engaging responses based on provided images and dialogue context. This task demands the optimal utilization of both textual and visual information. Early attempts at image-grounded dialogue generation \cite{ref13, ref12} were similar to visual dialogue, wherein the conversation topic revolves around a given image. However, recent research has shifted towards open-domain chitchat, utilizing the image to supplement conversation rather than dictate it.

The most elementary method, as delineated in \cite{ref20}, concatenates encoded image and text vectors before feeding them into the decoder to generate a response. Despite its simplicity, this approach neglects to capture the intricate relationship between visual and textual modalities, resulting in subpar responses. To overcome this limitation, researchers devised more sophisticated methods to more effectively incorporate visual information and bridge the modality gap. For example, Liang \textit{et al.} \cite{ref9} fed the encoded image information and embedded text into an autoregressive model \cite{ref21} to generate responses. During the decoding stage, the masked concept prediction objective was employed to assist the model in aligning multimodal representations. Similarly, Shen \textit{et al.} \cite{ref16} used a more complex model structure, passing the encoded image vector and text vector through a co-attention encoder and a cascade decoder to bridge the modality gap. Nevertheless, these methods still struggle with the modality gap problem due to the lack of alignment between different modality vectors.

Our proposed framework advances these existing methods and more effectively addresses the modality gap problem by incorporating a two-stage learning strategy. This design ensures a more thorough integration of visual and textual information, yielding more coherent and contextually appropriate responses in image-grounded dialogue generation.

\subsection{Zero-Resource Learning}
Zero-resource learning refers to scenarios where no training data exists for a specific task or domain. This scenario has been explored in various works, including Zero Resource Neural Machine Translation \cite{ref22, ref23}, Zero Resource Speech Challenge \cite{ref24, ref25}, and Zero Resource Dialogue Generation \cite{ref15, ref26}.

Li \textit{et al.} \cite{ref15} introduced a double latent variable model and a variational learning approach to investigate knowledge-grounded dialogue generation within a zero-resource scenario. Since then, the majority of dialogue generation work has been founded on this method, encompassing zero-resource image-grounded dialogue generation. For example, Yang \textit{et al.} \cite{ref8} hypothesized the presence of a latent image variable underpinning the textual dialogues and endeavored to recover it using text-to-image generation techniques. However, the text-to-image generation technique requires some annotated context-image-response data, which remains unavailable in zero-resource scenarios. Liang \textit{et al.} \cite{ref9} and Shen \textit{et al.} \cite{ref16} employed retrieval methods to obtain images corresponding to dialogue context, excluding the availability of context-image-response triples obtained from crowdsourcing for training purposes. The primary objective of these methods is to enhance dialogue generation by incorporating retrieved images. Although these methods have made significant progress, they still require annotated data for fine-tuning before effectively applying them to new domains.

Our proposed framework addresses the challenges inherent in a fully zero-resource scenario for image-grounded dialogue generation. By incorporating the learning strategy, our model effectively aligns multimodal vectors without relying on the annotated triples. This approach enables our model to generalize well to new domains where no labeled data is available for fine-tuning, making it a more practical solution for real-world applications.

\section{Methodology}
\subsection{Task Formalization and Model Overview}
The task of image-grounded dialogue generation can be defined as: given a dialogue context $C$ and an associated set of the top $k$ most relevant images $I_{1:k}$, the goal is to generate a coherent and contextually appropriate response $R$. The dialogue context $C$ consists of a sequence of tokens $c_1, c_2, \dots, c_n$, where $c_i$ represents the $i^{th}$ token in the conversation.

In a zero-resource scenario, the objective is to learn a text-image matching module $P(I_{1:k}|C)$, which provides a probability distribution over image sets, and a generative model $P(R|C, I_{1:k})$ capable of handling the set of top $k$ images collectively. By combining these two components, we obtain the following equation for a zero-resource image-grounded dialogue generation model $P(R|C)$:
\begin{equation}
	P(R|C) = P(R, I_{1:k}|C) = P(R|C, I_{1:k})P(I_{1:k}|C)
\end{equation}
The equality holds because the generative model can effectively utilize the information derived from the set of top $k$ images, and the text-image matching module is capable of providing a probability distribution over image sets.

Our proposed ZRIGF consists of four main components: a text-image matching module, a text-assisted masked image modeling module, a multimodal fusion module, and an information transfer module. These components are merged into our two-stage learning strategy: contrastive pre-training and generative pre-training.

\begin{figure*}
	\centering
	\includegraphics[scale=0.5]{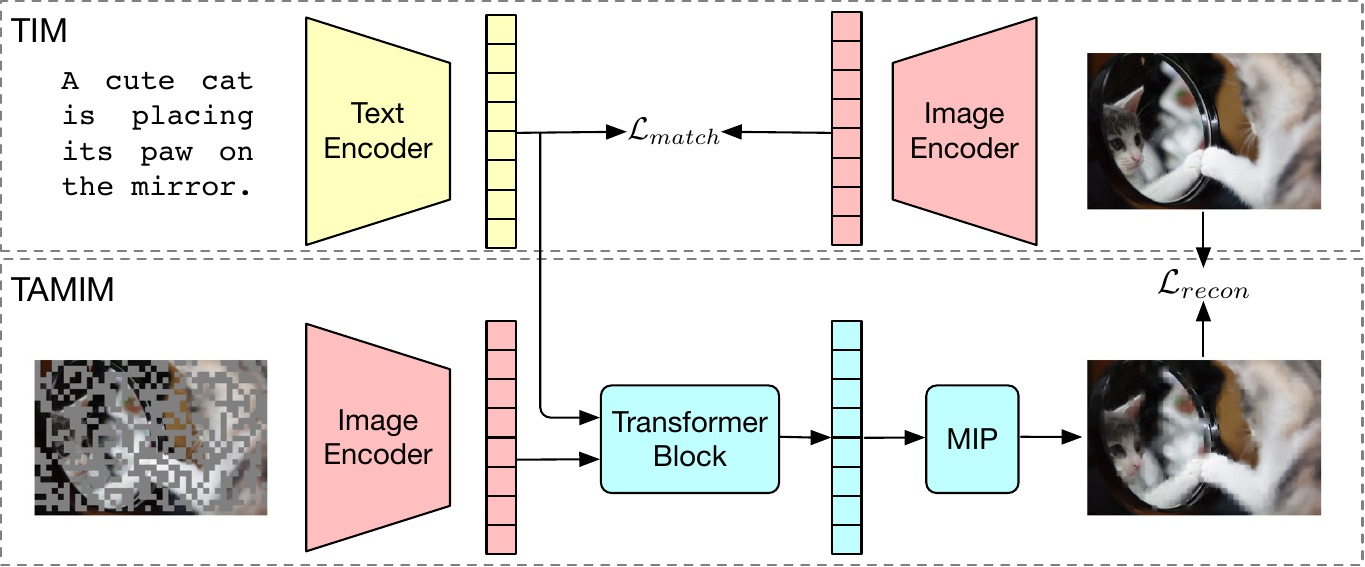}
	\caption{Overview of the contrastive pre-training stage.}
	\label{fig_con}
\end{figure*}

\subsection{Contrastive Pre-training}
As shown in Figure \ref{fig_con}, the aim of contrastive pre-training is to learn a text-image matching module $P(I_{1:k}|C)$ capable of providing a set of relevant images when given a dialogue context $C$. To achieve this, we utilize two modules: a text-image matching module and a text-assisted masked image modeling module. These modules are trained on a large corpus of annotated images and text pairs.

\subsubsection{Text-Image Matching Module}
The text-image matching (TIM) module takes an image-text pair as input and produces a similarity score indicating the extent of their correspondence. The input image is first processed by an image encoder $E_I$, initialized with a pre-trained ViT \cite{ref27}, to obtain the image hidden state $\mathbf{h}_I$. Then, we pool the model to acquire its image representation $\bar{\mathbf{h}}_I$ by simply taking $\mathbf{h}_I$ corresponding to the first token. Similarly, the input text is processed by a text encoder $E_T$, which is initialized with a pre-trained BART Encoder \cite{ref28}, to obtain its text representation $\bar{\mathbf{h}}_T$. We then compute the cosine similarity between $\bar{\mathbf{h}}_I$ and $\bar{\mathbf{h}}_T$ as follows:
\begin{equation}
	s_{IT} =  \cos(\bar{\mathbf{h}}_I,\bar{\mathbf{h}}_T) = \frac{\bar{\mathbf{h}}_I^T \bar{\mathbf{h}}_T}{\Vert\bar{\mathbf{h}}_I\Vert \Vert\bar{\mathbf{h}}_T\Vert}
	\label{eq2}
\end{equation}

We train this module using the CLIP loss \cite{ref29}, which aims to learn a joint embedding space for images and texts by maximizing the similarity between corresponding image-text pairs (i.e., positive pairs) and minimizing the similarity between non-matching image-text pairs (i.e., negative pairs). The loss function is defined as follows:
\begin{equation}
\begin{aligned}
	\mathcal{L}_{match} = -\mathbb{E}_{(I,T) \sim P(I,T)} \Big[&\log \frac{\exp(s_{IT}/\tau)}{\sum_{T' \sim P(T)} \exp(s_{IT'}/\tau)} \\
	+ \  &\log \frac{\exp(s_{IT}/\tau)}{\sum_{I' \sim P(I)} \exp(s_{I'T}/\tau)}\Big]
\end{aligned}
\label{eq_match}
\end{equation}
where $\tau$ represents a learnable temperature parameter controlling the distribution's sharpness, $P(I,T)$ is the joint distribution of images and texts, $P(I)$ is the marginal distribution of images, and $P(T)$ is the marginal distribution of texts.
By minimizing this loss function, we can align different modality vectors in the unified encoded vector space during the encoding process. This alignment can help bridge the modality gap and facilitate multimodal fusion in subsequent stages.

\subsubsection{Text-Assisted Masked Image Modeling Module}
While the alignment of different modality vectors is advantageous for multimodal fusion, it may also result in a loss of pre-training information due to projection. To address this issue, we employ a text-assisted masked image modeling (TAMIM) module during the contrastive pre-training stage. The TAMIM module takes a masked image and its corresponding text as input, in which a random portion of the image has been replaced with a mask token. The module's objective is to reconstruct the masked portion of the image using the unmasked regions, supplemented by the text representation.

To achieve this, we first obtain the masked image vector $\mathbf{h}_{I'}$ using the image encoder $E_I$. We then integrate the text vector $\mathbf{h}_T$ and the masked image vector $\mathbf{h}_{I'}$ through a transformer block $TB$ \cite{ref4}, consisting of multi-head attention and a position-wise feed-forward network. Next, we feed the output vector into a masked image prediction head $MIP$, composed of a convolution layer followed by a pixel-shuffle operation \cite{ref30}. This process can be described as follows:
\begin{equation}
	\mathbf{h}_{I'} = E_I(\mathbf{v}_I \odot \mathbf{M})
\end{equation}
\begin{equation}
	\hat{\mathbf{v}}_I = MIP(TB_I(\mathbf{h}_{I'}, \mathbf{h}_T, \mathbf{h}_T))
\end{equation}
where $\mathbf{v}_I$ denotes the original patch vector, $\hat{\mathbf{v}}_I$ represents the reconstructed patch vector, $\mathbf{M}$ is a binary mask with dimensions identical to $\mathbf{v}_I$, and $\odot$ indicates element-wise multiplication.

Afterward, we compute the reconstruction loss between the original patch vector $\mathbf{v}_I$ and the reconstructed patch vector $\hat{\mathbf{v}}_I$ using the mean absolute error (MAE) loss:
\begin{equation}
	\mathcal{L}_{recon} = \frac{1}{N_M} \sum_{i=1}^{N_M} |\mathbf{v}_I^{(i)} - \hat{\mathbf{v}}_I^{(i)}|
\end{equation}
where $N_M$ is the total number of masked image patches.
By minimizing this loss function, the TAMIM module can accurately reconstruct the masked portions of the image using the unmasked regions and the text representation. This can help preserve more pre-training visual information and further enhance the alignment between different modality vectors.

The overall loss function for the contrastive pre-training stage is a linear combination of the text-image matching loss and the text-assisted masked image modeling loss:
\begin{equation}
	\mathcal{L}_{con\_pre} = \mathcal{L}_{match} + \lambda_1 \mathcal{L}_{recon}
	\label{eq_con}
\end{equation}
where $\lambda$ represents a hyperparameter controlling the trade-off between the two objectives.

\begin{figure*}
	\centering
	\includegraphics[scale=0.5]{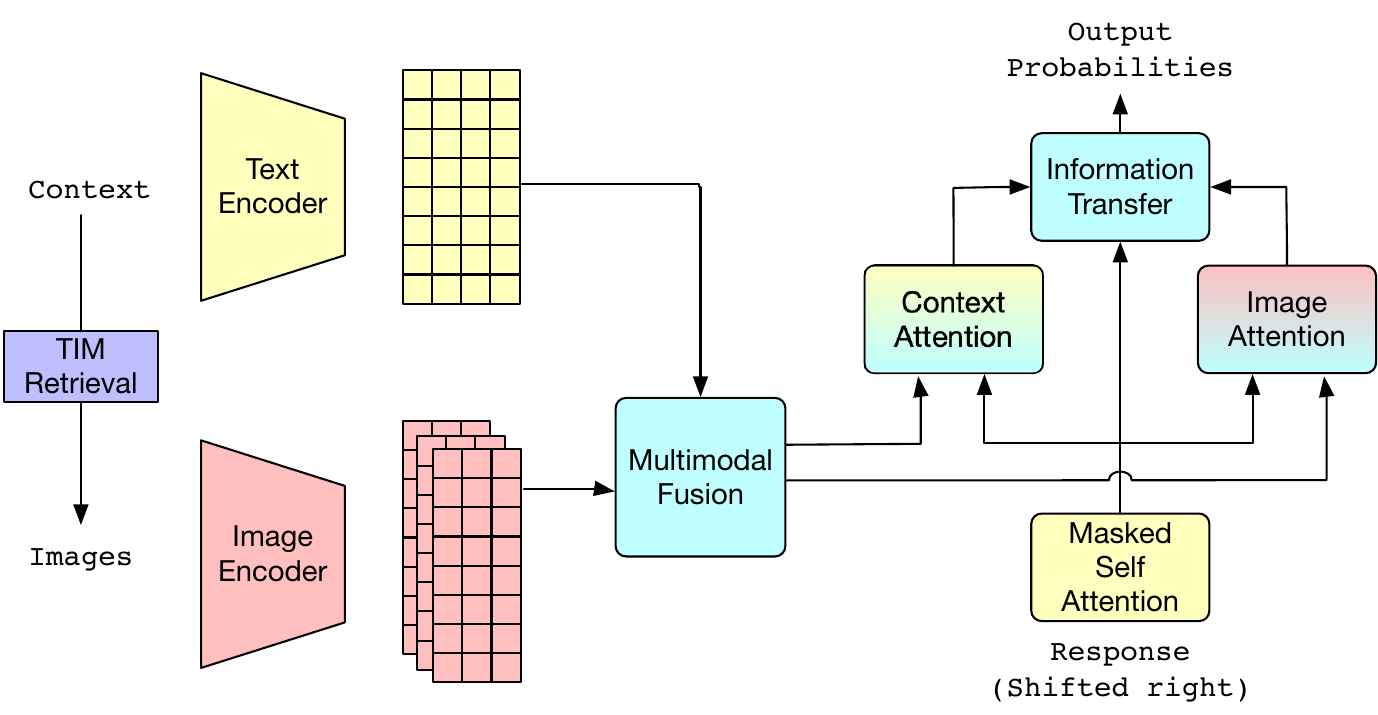}
	\caption{Overview of the generative pre-training stage.}
	\label{fig_gen}
\end{figure*}

\subsection{Generative Pre-Training}
As depicted in Figure \ref{fig_gen}, the generative pre-training stage's goal is to learn a generative model $P(R|C, I_{1:k})$ skilled in generating coherent and contextually appropriate responses when given a dialogue context $C$ and an associated set of images $I_{1:k}$. To accomplish this, we use a multimodal fusion module and an information transfer module. These modules are trained on a corpus of annotated context-response pairs, where the relevant images are obtained utilizing the TIM module. During training, the context and response are concatenated to obtain the relevant images, whereas, during testing, only the context is used.

\subsubsection{Multimodal Fusion Module}
The multimodal fusion (MF) module takes the dialogue context hidden state $\mathbf{h}_C$ and the top $k$ image hidden states $\mathbf{h}_{I_1}, \dots, \mathbf{h}_{I_k}$ as input, striving to effectively fuse multimodal information. Before achieving this, we initially apply an objective to ensure the encoders preserve the information contained in the TIM module. This objective aligns $\mathbf{h}_C$ and $\mathbf{h}_{I_1}, \dots, \mathbf{h}_{I_k}$ using binary cross-entropy loss:
\begin{equation}
	\mathcal{L}_{preser} = -\frac{1}{m} \sum_{i=1}^{n} \sum_{j=1}^{m} \left[ y_{ij} \log \sigma(s_{ij}) + (1 - y_{ij}) \log (1 - \sigma(s_{ij})) \right]
\end{equation}
where $n$ represents the number of dialogue contexts, $m$ represents the number of images, $y_{ij}$ is an element with $y_{ij} = 1$ if the $i^{th}$ context and the $j^{th}$ image form a matched pair, and $y_{ij} = 0$ otherwise, $\sigma(\cdot)$ is a logistic sigmoid function.

Then, we employ an attention mechanism to weigh the significance of each image representation concerning the dialogue context and perform the weighted sum over the image hidden state:
\begin{equation}
\alpha_i = \frac{\exp(s_{I_iC})}{\sum_{j=1}^{k}\exp(s_{I_jC})}, \quad \mathbf{h}_{I_{att}} = \sum_{i=1}^{k} \alpha_i \mathbf{h}_{I_i}
\end{equation}
This makes the fused image hidden state contain the beneficial information of $k$ images, while ignoring their noise information.

Next, we use two transformer blocks to fuse text and image hidden states separately:
\begin{equation}
	\mathbf{h}_{I_C} = TB_I(\mathbf{h}_{I_{att}}, \mathbf{h}_C, \mathbf{h}_C)
\end{equation}
\begin{equation}
	\mathbf{h}_{C_I} = TB_C(\mathbf{h}_C, \mathbf{h}_{I_{att}}, \mathbf{h}_{I_{att}})
\end{equation}

\subsubsection{Information Transfer Module}
When given the first $t-1$ response tokens $R_{1:t-1}$, the information transfer (IT) module predicts the probability of the $t^{th}$ token based on the fused dialogue context hidden state $\mathbf{h}_{C_I}$ and the fused image hidden state $\mathbf{h}_{I_C}$. This module is initialized with a pre-trained BART decoder, and the response hidden state $\mathbf{h}_R$ can be obtained from $R_{1:t-1}$ through its masked self-attention module. We then incorporate $\mathbf{h}_{C_I}$ and $\mathbf{h}_{I_C}$ into $R_{1:t-1}$ by the cross-attention module of the BART decoder to obtain $\mathbf{h}_{CR}$ and $\mathbf{h}_{IR}$, respectively. Next, an information transfer mechanism is employed to integrate all the information to obtain the predicted representation of $\hat{\mathbf{h}}_R$:
\begin{equation}
	\gamma_* = \sigma(\mathbf{W}_*[\mathbf{h}_R;\mathbf{h}_{*R}] + b_*)
\end{equation}
\begin{equation}
	\hat{\mathbf{h}}_R = FFN(\mathbf{W}_R[\mathbf{h}_R;\gamma_C \mathbf{h}_{CR};\gamma_I \mathbf{h}_{IR}] + b_R)
\end{equation}
where $_*$ here is denoted as $_C$ or $_I$, $\mathbf{W}_*$ and $b_*$ are learnable parameters, and $FFN$ is a position-wise feed-forward network. The information transfer mechanism selectively incorporates the fused dialogue context and image information into the response representation based on the context and image attentions $\gamma_C$ and $\gamma_I$, respectively.

Finally, the probability distribution of the $t^{th}$ token is obtained by applying a linear transformation followed by a softmax activation function $\eta(\cdot)$ to the corresponding $\hat{\mathbf{h}}_R$ at position $t-1$:
\begin{equation}
	P(R_t|R_{1:t-1},C,I_{1:k}) = \eta(\mathbf{W}_P \hat{\mathbf{h}}_{R_{t-1}} + b_P)
	\label{eq14}
\end{equation}
where the weight of $\mathbf{W}_P$ comes from the word embedding. The primary training objective of $P(R|C, I_{1:k})$ is to minimize the negative log-likelihood between the predicted probability distribution and the ground-truth response tokens:
\begin{equation}
\begin{aligned}
	\mathcal{L}_{gen} &= -\log P(R|C, I_{1:k})\\
	&= -\sum_t \log P(R_t|R_{1:t-1},C,I_{1:k})
\end{aligned}
\end{equation}

The overall loss function for the generative pre-training stage is a linear combination of the generative loss and the information preservation loss:
\begin{equation}
	\mathcal{L}_{gen\_pre} = \mathcal{L}_{gen} + \lambda_2 \mathcal{L}_{preser}
	\label{eq_gen}
\end{equation}

\subsection{Zero-Resource Learning Detail}
\subsubsection{Training}
During the zero-resource training stage, we initially train the contrastive pre-trained text-image matching module according to Equation \eqref{eq_con}, and subsequently train the generative pre-trained dialogue model as per with Equation \eqref{eq_gen}. ZRIGF is trained without explicit context-image-response pairs, as it leverages the learned representations of the individual components to enable effective dialogue generation.

Throughout the generative pre-training stage, the encoder undergoes alterations with each training iteration, leading to corresponding modifications in the TIM module. This results in a complex process of image retrieval based on the evolving TIM module. To enhance efficiency, we uniformly retrieve all relevant images after the contrastive pre-training stage.

\subsubsection{Inference}
For fully zero-resource scenarios in new domains, we simply use the pre-trained ZRIGF without any additional fine-tuning. During the zero-resource inference stage, the model first selects the top $k$ most relevant images $I_{1:k}$ for a given dialogue context $C$ using the TIM module $P(I_{1:k}|C)$ according to Equation \eqref{eq2}. Then, the generative model $P(R|C, I_{1:k})$ is utilized to generate a contextually appropriate response $R$ according to Equation \eqref{eq14}. As the model has been pre-trained to effectively fuse multimodal information and transfer this information into the generated response, it can generate coherent and contextually appropriate responses that are grounded in the provided images.

\section{Experiments}
\subsection{Datasets}
The Microsoft COCO 2017 dataset \cite{ref31} is utilized for contrastive pre-training to learn the text-image matching module and the text-assisted masked image modeling module. This dataset is segmented into 591K training samples and 25K validation samples. It comprises 123,287 images, each accompanied by five captions, providing an abundant source of image-text pairs for training both the TIM module and the TAMIM module.

The Reddit Conversation dataset \cite{ref19} is leveraged for generative pre-training, in which we train our model to generate contextually appropriate responses given a dialogue context. This dataset comprises over 15 million dialogues extracted from the Reddit platform, providing a large-scale and diverse collection of context-response pairs. The version used in this study, as released by \cite{ref8}, has been preprocessed to retain 1M dialogues for training, 20K dialogues for validation, and 20K dialogues for testing. During the generative pre-training, we utilize the TIM module learned from the contrastive pre-training stage to obtain the relevant images for the given dialogue contexts and responses. In addition to the images found in the COCO dataset, an extra 500K images from the Open Images dataset \cite{ref32} are employed as additional image indices.

The Image-Chat dataset \cite{ref12} is used for evaluating the performance of ZRIGF in zero-resource scenarios. This dataset is divided into 186,782 training examples, 5,000 validation examples, and 9,997 testing examples. It is specifically constructed for image-grounded dialogue generation tasks and comprises over 100,000 human-human dialogues grounded in images, with each dialogue consisting of a context, a relevant image, and a response. In the fully zero-resource scenario, we use only its test set for validation.

\subsection{Implementation Details}
We implement our ZRIGF based on the Hugging Face Transformers library \cite{ref33}. For the image encoder $E_I$, we initialize it with the ViT-Base model and use a patch size of 16. The text encoder $E_T$ and decoder are initialized with the BART-Large model. The image and text dimension are transformed by linear transformations. We employ the AdamW optimizer \cite{ref34} with a learning rate of $2e^{-5}$ for both contrastive and generative pre-training stages.

During the contrastive pre-training stage, the text encoder and decoder are frozen, and a batch size of 128 is employed. We train the model for 20 epochs, with a weight decay of 0.05 and a linear learning rate warm-up over the first $10\%$ of the total steps. We set the temperature parameter $\tau$ in Equation \eqref{eq_match} to 0.07, and the trade-off hyperparameter $\lambda_1$ in Equation \eqref{eq_con} to 0.2. For the TAMIM module, we use a random masking strategy with a patch size of $32 \times 32$ and a mask ratio of 0.4.

For the generative pre-training stage, we use a batch size of 32 and train the model for 10 epochs with a label smoothing factor of 0.1. In Equation \eqref{eq_gen}, the hyperparameter $\lambda_2$ is set to 0.1. The number of top relevant images $k$ is set to 3. Response tokens are generated using beam search with a beam size of 3.

\subsection{Baseline Models}
To evaluate the performance of our ZRIGF, we compare it against the following baseline models: (1) \textbf{Seq2Seq} \cite{ref35}, which is a standard sequence-to-sequence architecture consisting of an encoder and a decoder with long short-term memory (LSTM) cells; (2) \textbf{BART} \cite{ref28}, specifically the large BART model, which is a pre-training sequence-to-sequence model that employs a Transformer architecture and has demonstrated strong performance in various natural language generation tasks; (3) \textbf{ImgVAE} \cite{ref8}, which is an image-grounded dialogue generation model that leverages a variational autoencoder to incorporate visual information; (4) \textbf{Maria} \cite{ref9}, which is a neural conversation agent that can generate responses based on visual world experiences retrieved from a large-scale image index; and (5) \textbf{ChatGPT} \cite{ref11}, which is a state-of-the-art conversational AI model based on the GPT-3.5 and GPT-4 series of Large Language Models (LLMs), trained to follow an instruction in a prompt and provide a detailed response. Among them, ImgVAE and Maria are multimodal models, while all others are unimodal text models.

We have implemented all methods with the exception of ImgVAE, for which we directly utilized the results from \cite{ref8}. In the case of ChatGPT, we used GPT-3.5 and employed the following prompt to predict the conversation: \begin{quote} As a dialogue model, I would like you to predict the conversation, utilizing </s> as the conversation separator. Please provide the response only. \end{quote}

\subsection{Evaluation Metrics}
\subsubsection{Automatic Metrics}
To evaluate the performance of our proposed ZRIGF, we use several widely adopted automatic evaluation metrics. \textbf{Perplexity} (PPL) is used to assess the high-level general quality of the generation model, with a relatively lower Perplexity value indicating a more fluent response. Traditional metrics such as \textbf{BLEU-1} \cite{ref36} and \textbf{ROUGE-L} \cite{ref37} are used to compare the generated responses with reference human-generated responses, considering aspects such as 1-gram overlap and the longest common subsequence. Embedding-based metrics \cite{ref38} such as \textbf{Average}, \textbf{Extrema} and \textbf{Greedy} are used to compute the word embedding similarity between the generated response and the ground truth, based on cosine similarities. Diversity metrics \cite{ref39} such as \textbf{Dis-1} and \textbf{Dis-2} are used to report the degree of diversity in generated responses, measuring the ratio of unique uni/bi-grams over all uni/bi-grams.

\subsubsection{Human Evaluation}
For a more comprehensive evaluation, we conduct a human evaluation to assess the quality of the generated responses from different aspects. Specifically, we invite three human evaluators to rate 100 randomly generated responses in the test set on a scale from 1 to 5 in terms of \textbf{relevance}, \textbf{informativeness}, and \textbf{fluency}. Relevance evaluates how well the generated response relates to the given context and image, informativeness measures the amount of new and useful information provided by the response, and fluency assesses the grammaticality and readability of the generated text. The mean ratings are adopted as measures, and Fleiss' \textbf{Kappa} \cite{ref40} statistic is applied to evaluate the concordance among evaluators.

\begin{table*}
  \caption{Assessment of automated metrics: $\dag$ denotes a zero-resource scenario without annotated images, while $\ddag$ indicates a fully zero-resource scenario. Bold font highlights the best performance in each column.}
  \label{tab_automatic}
  \centering
  \begin{tabular}{cc *{8}{S}}
    \toprule
    {Task} & {Methods} & {PPL} & {BLEU-1} & {ROUGE-L} & {Average} & {Extrema} & {Greedy} & \ \  {Dis-1} & {Dis-2} \\
    \midrule
    \multirow{6}{*}{\makecell{Reddit\\ Conversation}}
    & Seq2Seq & 77.27 & 12.21 & 10.81 & 78.38 & 40.06 & 62.64 & 0.53 & 1.96 \\
    & BART & 44.73 & 13.51 & 12.50 & 80.21 & 41.63 & 63.72 & 4.17 & 16.98 \\
    & ChatGPT$^\ddag$ & {-} & 11.62 & 11.29 & \textbf{82.39} & 37.48 & 62.05 & 5.28 & \textbf{38.63} \\
    \cmidrule(lr){2-10}
    & ImgVAE & 72.06 & 12.58 & 12.05 & 79.95 & 42.38 & 63.55 & 1.52 & 6.34 \\
    & Maria & 56.23 & 14.10 & 12.66 & 81.76 & 43.04 & 63.98 & 4.83 & 22.87 \\
    \cmidrule(lr){2-10}
    & ZRIGF & \textbf{36.21} & \textbf{16.06} & \textbf{14.51} & 82.27 & \textbf{43.79} & \textbf{64.53} & \ \ \textbf{5.79} & 26.57 \\
    \midrule
    \multirow{9}{*}{Image-Chat}
    & Seq2Seq$^\dag$ & 50.82 & 11.34 & 13.65 & 82.95 & 47.45 & 65.67 & 1.28 & 7.80 \\
    & BART$^\dag$ & 37.26 & 13.41 & 14.24 & 84.48 & 48.57 & 66.49 & 2.44 & 15.79 \\
    & ChatGPT$^\ddag$ & {-} & 10.77 & 11.62 & 86.17 & 43.02 & 64.66 & \ \ \textbf{5.32} & \textbf{37.77} \\
    \cmidrule(lr){2-10}
    & ImgVAE & 41.94 & 16.07 & 15.98 & 85.81 & 49.59 & 67.44 & 1.68 & 7.22 \\
    & Maria$^\dag$ & 37.49 & 14.74 & 14.59 & 85.72 & 50.58 & 66.89 & 2.57 & 11.99 \\
    & Maria$^\ddag$ & 135.49 & 11.75 & 12.13 & 83.51 & 45.57 & 64.48 & 1.89 & 7.32 \\
    \cmidrule(lr){2-10}
    & ZRIGF & \textbf{29.82} & 16.86 & \textbf{17.21} & \textbf{86.30} & \textbf{51.41} & \textbf{68.56} & 2.59 & 10.62 \\
    & ZRIGF$^\dag$ & 30.58 & \textbf{17.29} & 17.03 & 86.00 & 50.99 & 68.05 & 4.28 & 24.10 \\
    & 1/4 Data$^\dag$ & 35.41 & 16.35 & 16.59 & 85.75 & 49.95 & 67.20 & 4.61 & 22.66 \\
    & 1/8 Data$^\dag$ & 40.07 & 16.01 & 16.08 & 84.91 & 48.74 & 66.32 & 4.31 & 19.17 \\
    & Zero Data$^\ddag$ & 105.12 & 15.17 & 15.13 & 84.52 & 45.95 & 65.70 & 5.25 & 29.38 \\ 
    \bottomrule
  \end{tabular}
\end{table*}

\section{Results and Discussion}
\subsection{Automatic Evaluation Results}
Table \ref{tab_automatic} presents the results of the automatic evaluation metrics for the proposed ZRIGF and the baseline models on the Reddit Conversation and Image-Chat datasets. For the Reddit Conversation dataset, we observe that our ZRIGF outperforms all the baseline models in terms of PPL, BLEU-1, ROUGE-L, Extrema, Greedy, and Dis-1 metrics, demonstrating the effectiveness of the proposed multimodal architecture and the learning strategy in generating contextually appropriate responses. ChatGPT outperforms ZRIGF on the Average and Dis-2 metrics. This is because ChatGPT is based on LLM, which enables it to generate more diverse responses.

For the Image-Chat dataset, we evaluate our ZRIGF in three scenarios: a scenario with annotated images, a zero-resource scenario without annotated images ($\dag$), and a fully zero-resource scenario without any training data ($\ddag$). We also report the results of our ZRIGF when trained with 1/4 and 1/8 of the training data to examine its robustness under data scarcity. We can see that ZRIGF$^\dag$ achieves the best performance on most metrics compared to baseline models, indicating its effectiveness in generating informative and coherent responses that are grounded on images. The results also show that ZRIGF can handle the zero-resource scenario well, as ZRIGF$^\dag$ performs competitively with ZRIGF and its diversity has been significantly enhanced. Notably, even when trained with only 1/4 of the available training data in the second scenario, ZRIGF still surpasses ImgVAE, which used all available training data. Moreover, ZRIGF shows remarkable robustness and generalization ability in different data settings, as its performance does not degrade significantly when reducing the amount of training data or using zero data. In contrast, Maria suffers from a severe performance drop when switching from full training data to zero training data. ChatGPT performs well in terms of Dis-1 and Dis-2 metrics, but it lags behind ZRIGF$^\ddag$ in terms of BLEU-1 and ROUGE-L metrics, suggesting that it may generate responses that are diverse but not specific or relevant to the given context.

\begin{table}
  \caption{Human evaluation outcomes for the Image-Chat dataset in a fully zero-resource scenario.}
  \label{tab_human}
  \centering
  \begin{tabular}{c *{4}{S}}
    \toprule
    {Methods} & {Relevance} & {Informativeness} & {Fluency} & {Kappa} \\
    \midrule
    Maria & 2.26 & 1.56 & 3.37 & 0.41 \\
    ChatGPT & 3.38 & \ \ \textbf{3.02} & \ \ \textbf{4.46} & 0.48 \\
    ZRIGF & \ \ \textbf{3.74} & 2.97 & 4.23 & 0.45 \\
    \bottomrule
  \end{tabular}
\end{table}

\subsection{Human Evaluation Results}
Table \ref{tab_human} shows the results of the human evaluation for the Image-Chat task in a fully zero-resource scenario. We compare ZRIGF with Maria and ChatGPT models. We can see that ZRIGF obtains the highest rating in terms of relevance, indicating that it can generate responses that are closely related to the given context and image. ChatGPT achieves the highest rating in terms of informativeness and fluency, suggesting that it can generate responses that are rich in content and grammatically correct. However, ChatGPT also receives a lower rating in terms of relevance than ZRIGF, implying that it may generate responses that are informative but not specific or consistent with the given context and image. It is possible that ChatGPT is based on a text-only LLM, which does not incorporate multimodal information. ZRIGF achieves results similar to ChatGPT in terms of informativeness, thanks to the incorporation of multimodal information. Maria obtains the lowest rating in all aspects, reflecting its poor performance in a fully zero-resource scenario.

\begin{figure*}[h]
	\centering
	\includegraphics[scale=0.5]{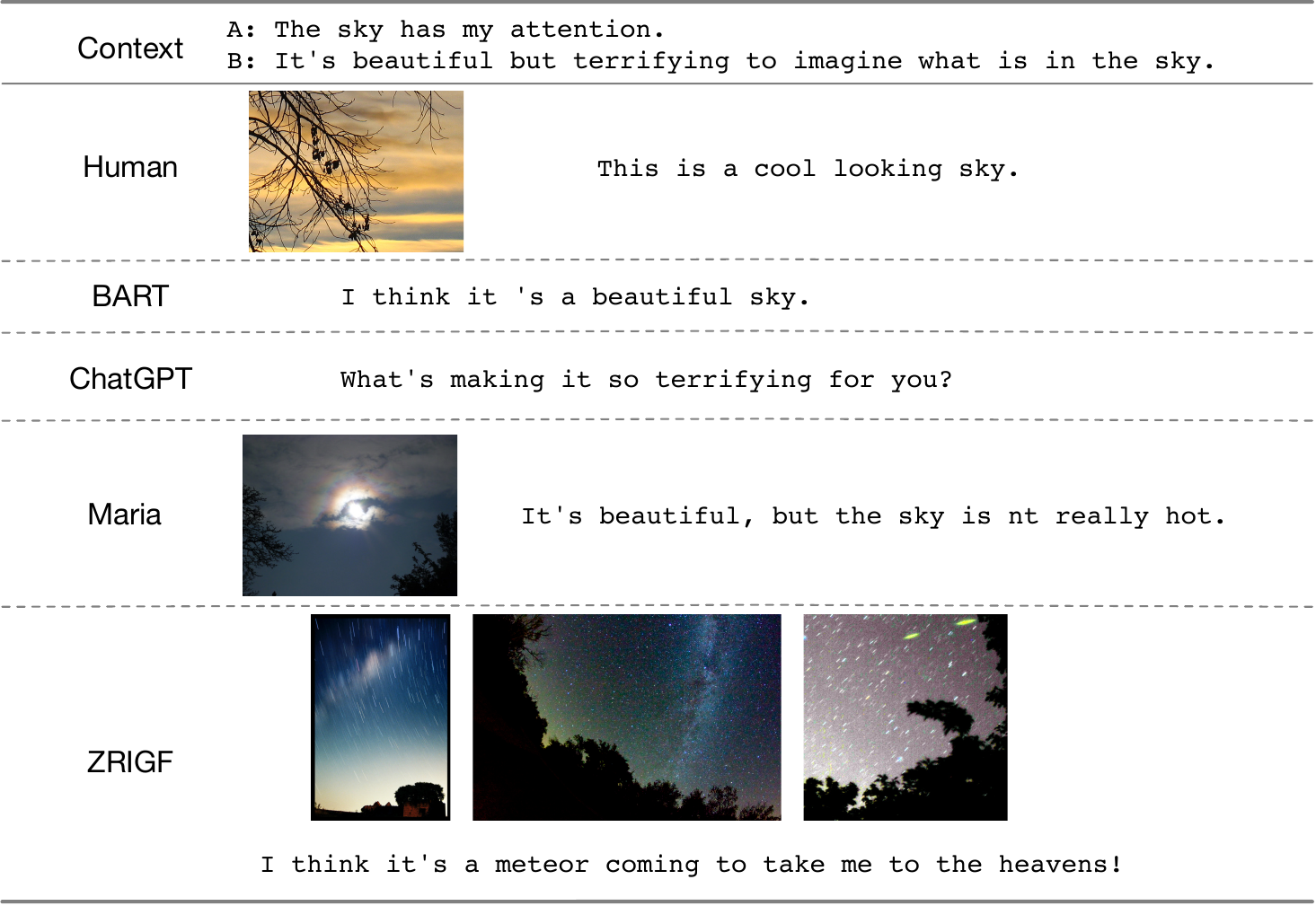}
	\caption{Case study on Image-chat test set in a fully zero-resource scenario. Maria and ZRIGF retrieve relevant images based on context and generate responses accordingly.}
	\label{fig_case}
\end{figure*}

\begin{table}
  \caption{Ablation study.}
  \label{tab_ablation}
  \centering
  \begin{tabular}{c *{4}{S}}
    \toprule
    \multirow{2}{*}{Methods}
    & \multicolumn{2}{c}{Reddit Conversation} & \multicolumn{2}{c}{Image-Chat (Zero Data)} \\
    \cmidrule(lr){2-3} \cmidrule(lr){4-5}
    & {BLEU-1} & {ROUGE-L} & {BLEU-1} & {ROUGE-L}\\
    \midrule
    ZRIGF & \textbf{16.06} & 14.51 & \textbf{15.17} & \textbf{15.13} \\
    \midrule
    -TIM & 15.19 & 13.95 & 14.77 & 15.02 \\
    -TAMIM & 13.68 & 11.33 & 12.33 & 13.31 \\
    -MF & 15.46 & \textbf{14.56} & 14.90 & 14.89 \\
    -IT & 15.30 & 13.89 & 14.17 & 14.50 \\
    \bottomrule
  \end{tabular}
\end{table}

\subsection{Ablation Study}
To analyze the impact of each module in ZRIGF, we conduct an ablation study by removing one module at a time and evaluating the performance on both Reddit Conversation and Image-Chat datasets. Table \ref{tab_ablation} shows the results of the ablation study in terms of BLEU-1 and ROUGE-L metrics. We observe that removing any module leads to a decrease in performance, indicating that each module contributes to the overall quality of the generated responses. Among the four modules, we find that removing TAMIM causes the most significant drop in performance, suggesting that TAMIM plays a vital role in preserving and realigning the multimodal features for dialogue generation. Removing IT also results in a noticeable decline in performance, especially in the fully zero-resource scenario, which implies that IT is critical to the generalization ability of ZRIGF. TIM and MF are relatively less essential but still beneficial for generating insightful responses based on multimodal representations.

We further explored the impact of the relevance of retrieved images on the generation performance, detailed in Appendix \ref{appendix_a}.

\subsection{Case Study}
To further illustrate the effectiveness of the proposed ZRIGF, we conduct a case study by analyzing the responses generated in Figure \ref{fig_case}. We compare our model to baseline models on the Image-Chat test set in a fully zero-resource scenario. As can be seen from the figure, ZRIGF effectively utilizes the visual information from retrieved images, generating more detailed and engaging responses. ChatGPT, benefiting from a powerful LLM, produces fluent and contextually relevant responses; however, its inability to incorporate visual information results in less engaging responses compared to ZRIGF. Maria generates a response somewhat related to the retrieved image, but it lacks relevance and accuracy. This case study demonstrates that ZRIGF successfully leverages the integrated multimodal information to generate more image-grounded and contextually appropriate responses.

\section{Conclusion}
In this paper, we have proposed ZRIGF, an innovative multimodal framework for zero-resource image-grounded dialogue generation. ZRIGF consists of four modules: a text-image matching module, a text-assisted masked image modeling module, a multimodal fusion module, and an information transfer module. ZRIGF implements a two-stage learning strategy that combines contrastive pre-training and generative pre-training to align multimodal features and generate insightful responses. We have evaluated ZRIGF on two publicly available dialogue datasets and demonstrated its effectiveness and robustness in generating cohesive and contextually pertinent responses that are grounded on images. We have also shown that ZRIGF can generalize well in the fully zero-resource scenario without requiring any training data. Our work contributes to the advancement of image-grounded dialogue systems and opens up new possibilities for multimodal dialogue generation in zero-resource scenarios.

\begin{acks}
This research is supported by the Natural Science Foundation of China (No. 62006034).
\end{acks}

\bibliographystyle{ACM-Reference-Format}
\balance
\bibliography{sample}

\appendix

\section{Analysis of Image Relevance}
\label{appendix_a}

\subsection{Impact of irrelevant Images}

\begin{table}[h]
  \caption{Performance comparison with the incorporation of irrelevant images.}
  \label{tab_irrelevant}
  \centering
  \begin{tabular}{c *{4}{S}}
    \toprule
    \multirow{2}{*}{Methods}
    & \multicolumn{2}{c}{Reddit Conversation} & \multicolumn{2}{c}{Image-Chat (Zero Data)} \\
    \cmidrule(lr){2-3} \cmidrule(lr){4-5}
    & {BLEU-1} & {ROUGE-L} & {BLEU-1} & {ROUGE-L}\\
    \midrule
    ZRIGF & \textbf{16.06} & \textbf{14.51} & \textbf{15.17} & \textbf{15.13} \\
	BART & 13.51 & 12.50 & 13.41 & 14.24 \\
	Random & 12.11 & 11.47 & 12.32 & 13.60 \\
    \bottomrule
  \end{tabular}
\end{table}

The impact of irrelevant images on generation performance is a critical factor in understanding the overall behavior and dependency of the ZRIGF model on visual information. To investigate this, we randomly replaced the top k relevant images with irrelevant ones taken from the Open Images dataset. Table \ref{tab_irrelevant} demonstrates the comparison between the performance of the ZRIGF model, BART, and a Random selection approach. The BART model can be considered as the case where no images are used. It can be observed that the BLEU-1 and ROUGE-L scores drop markedly for ZRIGF when irrelevant images are used, compared to the BART model, where no images are incorporated.

The results underline the ZRIGF's strong reliance on the quality of retrieved images, showing that the presence of irrelevant images has negative effects on response generation. This can be attributed to the model's design, which presumably leverages the visual context in generating responses. Therefore, the incorporation of irrelevant images likely confuses the model, leading to less coherent and contextually accurate output. This also demonstrates the model's ability to retrieve relevant images.

\subsection{Impact of Image Quantity}

\begin{table}[h]
  \caption{Performance comparison with various numbers of images.}
  \label{tab_quantity}
  \centering
  \begin{tabular}{c *{4}{S}}
    \toprule
    \multirow{2}{*}{Methods}
    & \multicolumn{2}{c}{Reddit Conversation} & \multicolumn{2}{c}{Image-Chat (Zero Data)} \\
    \cmidrule(lr){2-3} \cmidrule(lr){4-5}
    & {BLEU-1} & {ROUGE-L} & {BLEU-1} & {ROUGE-L}\\
    \midrule
	k=1 & 15.54 & 13.72 & 13.74 & 14.99 \\
	k=3 & \textbf{16.06} & \textbf{14.51} & \textbf{15.17} & \textbf{15.13} \\
	k=5 & 15.77 & 14.49 & 15.04 & 15.02 \\
    \bottomrule
  \end{tabular}
\end{table}

The role of hyper-parameter $k$, which controls the number of retrieved images, is a key aspect to consider in understanding the ZRIGF model's behavior. We conducted a series of experiments to study how varying the value of $k$ affects the generation performance. Specifically, we set $k$ to 1, 3, and 5 and analyzed the model's output with different quantities of retrieved images. Table \ref{tab_quantity} illustrates the performance of ZRIGF for different values of $k$. The best results were achieved when $k = 3$, indicating an optimal trade-off between retrieving more images and retrieving more relevant images. When $k$ is too small, ZRIGF may miss valuable images that could provide essential information for response generation. On the other hand, if $k$ is too large, the model might retrieve irrelevant or redundant images, leading to ambiguity and potential noise in the response generation process.

\end{document}